\documentclass[journal]{IEEEtran}
\usepackage[sort&compress,numbers]{natbib}
\usepackage{amsmath,amssymb,amsfonts}
\usepackage{algorithm}
\usepackage{algpseudocode}
\usepackage{graphicx}
\usepackage{textcomp}
\usepackage{blindtext}
\usepackage{subfiles}
\usepackage{standalone} 
\usepackage{tikz}
\usetikzlibrary{positioning, fit}
\usetikzlibrary{backgrounds}
\usepackage{pgfplots}
\usepackage{url}
\pgfplotsset{compat = newest}

\DeclareFontFamily{U}{rcjhbltx}{}
\DeclareFontShape{U}{rcjhbltx}{m}{n}{<->rcjhbltx}{}
\DeclareSymbolFont{hebrewletters}{U}{rcjhbltx}{m}{n}

\let\aleph\relax

\DeclareMathSymbol{\aleph}{\mathord}{hebrewletters}{39}

\def\BibTeX{{\rm B\kern-.05em{\sc i\kern-.025em b}\kern-.08em
    T\kern-.1667em\lower.7ex\hbox{E}\kern-.125emX}}

\begin{document}
\title{Motion-compensated recovery using diffeomorphic flow (DMoCo) for 5D cardiac MRI} 
\author{Joseph William Kettelkamp, Ludovica Romanin, PhD,  Sarv Priya, MD, and  Mathews Jacob, PhD,
\thanks{This work is supported by NIH grants: R01EB019961, R01AG067078, and R01AG087159. Joseph William Kettelkamp and Mathews Jacob are with University of Virginia (e-mail:{trz8sp,mjacob}@virginia.edu). Ludovica Romanin is with Siemens Healthineers International AG, Lausanne, Switzerland. Sarv Priya is with University of Wisconsin in Madison. The clinical data was acquired at the University of Iowa Hospitals and Clinics (UIHC).}
} 

\maketitle

\begin{abstract} We introduce an unsupervised motion-compensated image reconstruction algorithm for free-breathing and ungated 5D functional cardiac magnetic resonance imaging (MRI). The proposed algorithm models the 3D volumes at different cardiac and respiratory phase as deformed versions of a single static image template. The template as well as the parameters of the deformations are jointly estimated from the measured k-t space data. The main contribution of this work is a constrained representation of the deformations as a family of diffeomorphisms, parameterized by the motion phases. The deformation at a specific motion phase is derived as the integral of a parametric velocity tensor along a path connecting the reference template phase to the motion phase. The velocity tensor at different phases is represented using a low-rank model. The results show that the more constrained motion model offers improved recovery compared to current motion-resolved and motion-compensated algorithms for free-breathing and ungated 3D cardiac MRI data. 

\end{abstract}

\begin{IEEEkeywords}
diffeomorphism, low-rank, motion compensation, unsupervised, moco
\end{IEEEkeywords}

\section{Introduction}
\label{sec:introduction}

\IEEEPARstart Magnetic Resonance (MR) is the gold standard for evaluating cardiac function. Current clinical 2D cine acquisition strategy requires extensive preparatory scans for slice localization and requires breath hold, which many subject groups (e.g., children or patients with reduced tidal volume) cannot comply with. 3D free-breathing and ungated acquisition schemes that offer isotropic spatial resolution and eliminate the need for preparatory scans were recently introduced \cite{Liu2010,Pang2014,Coppo2014,Usman2017,Feng2017,Moghari2017,DiSopra2019,Rosenzweig2020,Kstner2020,Roy2024,Arshad2024}. 
These schemes estimate the respiratory and cardiac phases from the center of k space, dedicated k space navigators, or ECG signals. The k-space data are grouped into cardiac and respiratory bins using the above phase information, followed by joint reconstruction of the images using total variation or low-rank penalties. Although these motion-resolved approaches offer great promise, the main challenge is the associated long acquisition times to ensure good and even sampling of the bins. Although supervised deep learning (DL) approaches may offer improved trade-offs, the challenge is the lack of training data; unlike 2D settings, the acquisition of fully sampled images is not possible in our 5D setting. In addition, the extension of current DL methods is challenging because of the high memory demand in the 3D setting. Recently, patient-specific self-supervised deep learning methods were introduced 
\cite{Yoo2021Time-DependentMRI,qing,Spieker2023ICoNIK:K-Space,Huang2022NeuralImaging,Li2023LearningMatching,qingvariational} to reduce data demand in motion-resolved 4D MRI (two spatial dimensions along with cardiac and respiratory). These methods express the image at a specific cardiac and respiratory phase as a nonlinear function of the low-dimensional signals (e.g. cardiac and respiratory motion). In our 5D setting, a large 3D CNN is needed to represent the image volumes in each phase; training of this model using backpropagation will be associated with high memory demand.

Motion-compensated (MoCo) reconstruction schemes that explicitly model the deformations between the phases are significantly more data efficient than the above motion resolved strategies \cite{Odille2016JointQuantification,Munoz2022Self-supervisedAngiography,MoCo-MoDL,iMoCo}. Because they combine the information from the different motion phases, they are less sensitive to varying numbers of samples in each bin. The main drawback of these methods is the high computational complexity of image registration. Deep learning (DL) methods for estimating respiratory motion \cite{Munoz2022Self-supervisedAngiography} and unrolled DL methods that incorporate motion compensation have recently been introduced \cite{Pan2024UnrolledMRI}. These methods are challenging to apply directly to our 5D setting because of the extreme and uneven undersampling of the bins, the high memory demand, and the need for fully sampled reference data. An unsupervised DL approach called MoCo-SToRM \cite{moco-storm} was recently introduced for respiratory motion-compensated lung MRI. MoCo-SToRM models the volume in each respiratory phase as the deformed version of a static image template. The motion field at each phase is modeled as the output of a CNN in response to low-dimensional latent vectors that correspond to the phase. The latent vectors, the CNN parameters, and the image template are jointly estimated from the undersampled k-t space data of each subject, similar to self-supervised motion-resolved strategies\cite{Yoo2021Time-DependentMRI,qing,Spieker2023ICoNIK:K-Space,Huang2022NeuralImaging,Li2023LearningMatching,qingvariational,Chandrasekaran2025}. We note that MoCo-SToRM was originally developed for respiratory-compensated 4D MRI. The direct extension of the simple deformation model in MoCo-SToRM to our respiratory and cardiac motion-compensated setting can result in inaccurate deformation estimates, especially when the phases are heavily undersampled and the contrast-to-noise ratio is low.  Although approaches similar to MoCo-SToRM \cite{OSUCardiac,Kettelkamp2023} have recently been introduced for the 5D setting, they often require ferumoxytol infusion that offers an order of magnitude improvement in contrast to the noise ratio to obtain reliable results \cite{OSUCardiac}. Because ferumoxytol administration is not a viable option in many clinical settings, the development of methods for a low SNR and a highly accelerated setting is desirable. 

We introduce an unsupervised approach for the recovery of motion-compensated cardiac cine acquired using free breathing and ungated 3D-SSFP acquisitions. Similar to MoCo-SToRM \cite{Kettelkamp2023}, we represent the image in each phase (cardiac and respiratory) as the deformed version of a static image template. The main distinction of the proposed scheme with MoCo-SToRM is the more constrained low-rank diffeomorphic motion model. The deformation model in MoCo-SToRM does not guarantee that the deformation of the heart and surrounding tissues maintains integrity and connectivity and is devoid of intersections between different regions. We are motivated by diffeomorphic models used in medical image registration that represent the deformation between two images as an invertible function, so that both the function and its inverse are continuously differentiable \cite{voxelmorph,christensen}. These schemes express the deformation as the time integral of a static vector field, which guarantees the above properties. The main contribution of this work is a low-rank model for the compact joint representation of the family of diffeomorphisms. This work involves the generalization of current diffeomorphic models \cite{christensen}, which consider the registration of a pair of images, to the joint registration of a family of images. In particular, we model the deformation at each motion phase as the integral of a phase-dependent velocity tensor along a linear path connecting the template phase and the desired phase. We observe that the collection of the velocity tensors at different motion phases is significantly more low-rank than the original images and the deformation fields at different phases. The low-rank velocity representation thus offers a compact model of the diffeomorphisms, and hence the images. We additionally incorporate a regularization prior to encourage the diffeomorphisms to be independent of the integration path, which further reduces the degree of freedom. We estimate the velocity tensors and the static template from the measured k-space data in an unsupervised fashion. The coupling of the motion fields at different phases via the overlapping path integrals, the use of velocity constraints the deformations independent of the integration path, and the low-rank model make the deformation model more constrained than the unsupervised approach in \cite{moco-storm} and the supervised approach \cite{Munoz2022Self-supervisedAngiography}. The preliminary studies in the paper shows the feasibility of DMoCo to enable cardiac and respiratory resolved recovery of 5D cardiac MRI data from six minutes of radial free-breathing and ungated acquisition. The main objective of this work is on introduce the DMoCo framework and to show its feasibility in this setting; future work would focus on a more thorough clinical validation.  

\section{Background}
\subsection{Multichannel MRI Acquisition}
The multichannel acquisition of the data can be mathematically modeled as: 
\begin{equation}
    \mathbf b(t) = \mathcal A_{\mathbf k(t)} \left(\rho_t(\mathbf r)\right) + \mathbf n(t).
\end{equation}
Here, $\mathbf r = (x,y,z)\in \mathbb R^3$ denotes the spatial variable and $t$ denotes the time at which the k-space measurements $\mathbf b(t)$ were acquired. $\rho_t$ denotes the spatial volume at the time-instant $t$ and $\mathcal A_{\mathbf k(t)}$ denotes the multichannel Fourier transform corresponding to the k-space locations $\mathbf k(t)$. $\mathbf n(t)$ denotes the measurement noise of variance $\sigma^2$.

\subsection{Diffeomorphic Image Registration}
\label{background}

Given two $n$-dimensional image volumes $I_0, I_1\in \mathbb R^n \rightarrow \mathbb C$, image registration aims to find a deformation $\phi_1: \mathbb R^n \rightarrow \mathbb R^n$ such that $I_1( \mathbf r) = I_0(\phi_1(\mathbf r))$. A deformation may be evaluated as $\phi_1(\mathbf r) = \mathbf r + \boldsymbol u(\mathbf r)$, where $\boldsymbol u$ denotes the displacement. Although its inverse may be approximated as $\mathbf r = \phi_1(\mathbf r) - \boldsymbol v(\phi_1(\mathbf r))$, this is only as exact when $\boldsymbol u(\mathbf r) \rightarrow \boldsymbol v(\phi(\mathbf r))$; that is, for infinitesimal deformations that satisfy $\boldsymbol v\rightarrow 0$. One can compose several infinitesimal deformations $$
  \phi_N(\mathbf r) = (\mathbf r+\boldsymbol v_N(\mathbf r))~\circ~ \ldots  ~\circ~ (\mathbf r+\boldsymbol v_0(\mathbf r)),$$  
to obtain a large deformation that is invertible. This intuition has led to the representation of diffeomorphisms, which are often expressed as the endpoint of a flow. $\phi_{\delta}(\mathbf r):[0,1]\times\mathbb R^n \rightarrow \mathbb R^n$, associated with a smooth $n$-dimensional vector field $\boldsymbol \nu_{\delta}(\mathbf r): [0,1]\times\mathbb R^n \rightarrow \mathbb R^n$, specified by the ordinary differential equation \cite{beg,christensen} 
$$  \frac{d \phi_{\delta}(\mathbf r)}{d\delta} = \boldsymbol \nu_{\delta}\Big(\phi_{\delta}(\mathbf r)\Big); ~~~\phi_0(\mathbf r)=\mathbf r.
$$
The above ODE is solved as 
\begin{equation}
\label{ode}
    \phi_1(\mathbf r) =  \underbrace{\phi_0(\mathbf r)}_{\mathbf r} + \int_{0}^1 \boldsymbol \nu_{\delta}\Big(\phi_{\delta}(\mathbf r)\Big)~ d\delta
\end{equation}
\eqref{ode} can be inverted by running the ODE backwards with an initialization $\phi_{1}(\mathbf r)=\mathbf r$ as 
\begin{equation}
\label{odeback}
  \mathbf \phi_0\left( \mathbf{r} \right) =  \mathbf r  + \int_{1}^0 \boldsymbol \nu_{\delta}\Big(\phi_{\delta}(\mathbf r)\Big)~ d\delta  
\end{equation}

\section{Low-rank diffeomorphic Image Model}

The diffeomorphic flow-based motion compensated (DMoCo) approach represents each volume in the time series as a deformed version of a static template. We now describe the main aspects of the framework, which are (a) image representation, (b) diffeomorphic motion model using a parametric velocity representation, and (c) low-rank model for the velocities.
\subsection{Image representation}
We represent the image volume $\rho_t(\mathbf r)$ at the time instant $t$ as a function of the continuous phase variable $\boldsymbol \tau(t)\in \mathbb R^{2}$.
\begin{equation}
\label{latent}
    \boldsymbol \tau(t) = \begin{bmatrix}
    c(t)\\
    r(t)
    \end{bmatrix}
\end{equation}
Here, $c(t)$ and $r(t)$ denote the cardiac and respiratory phase at $t$. We assume that the images in the time series depend on the motion phase $\boldsymbol \tau(t)$.We represent the image volume at time $t$ specified by $\rho_t(\mathbf r)$ at the motion phase $\boldsymbol\tau(t)$ as the deformed version of a static template $\eta:\mathbb R^{3}\rightarrow \mathbb C$:
\begin{equation}\label{imagemodel}
    \rho_{t}(\mathbf r) = \eta\Big( \varphi_{\boldsymbol \tau(t)}(\mathbf r)\Big) .
\end{equation}
Here, $\varphi_{\boldsymbol \tau}(\mathbf r)$ is the deformation at the motion phase $\boldsymbol{\tau}$. 

\subsection{Parametric Diffeomorphic Motion Model}
The main distinction of DMoCo from MoCo-SToRM is the more constrained model used to represent $\varphi_{\boldsymbol \tau}(\mathbf r)$. Most diffeomorphic registration methods consider the registration of a pair of images. In this work, we consider the joint registration of images $\rho_t(\mathbf r)$ using a parametric family of diffeomorphisms, parameterized by $\boldsymbol \tau$. The diffeomorphisms are represented in terms of a parametric velocity tensor $\mathbf v_{\boldsymbol \tau}\Big(\frac{d\boldsymbol{\tau}}{dt},\mathbf r\Big)$ at the motion phase $\boldsymbol \tau$. The tensor $\mathbf v_{\boldsymbol \tau}$ can be thought of as the concatenation of two velocity fields. For example, the first component of the tensor denoted by $\mathbf v_{\boldsymbol \tau}(0,\mathbf r)$ is the tissue velocity in the phase $\boldsymbol \tau$, when the cardiac phase increases at the rate $dc/dt=1$. When both phases are increased at a speed  $\boldsymbol\tau'=\begin{bmatrix}dc/dt&dr/dt\end{bmatrix}^T$, we assume linearity and obtain the tissue velocity as
$$\mathbf v_{\boldsymbol\gamma}\Big(0,\mathbf r\Big)~ \frac{dc}{dt} + \mathbf v_{\boldsymbol\gamma}\Big(1,\mathbf r\Big)~ \frac{dr}{dt} = \Big\langle \mathbf v_{\boldsymbol\gamma}(\mathbf r),\frac{d\boldsymbol\tau}{dt}\Big\rangle.$$

We now consider a curve $\boldsymbol\gamma: [0,1] \rightarrow \mathcal C_{\boldsymbol \tau}$ in phase space (see Fig. \ref{Fig:art}) starting at $\mathbf 0$ and ending at $\boldsymbol \tau$ with $\boldsymbol\gamma(0)=\mathbf 0$ and $\boldsymbol\gamma(1)= \boldsymbol \tau$ to define the diffeomorphism $\varphi_{\boldsymbol\tau}(\mathbf r)$.  $\boldsymbol\gamma(\delta)$ are points on the curve, and $\delta$ is an arbitrary parameterization. For the curve $\mathcal C_{\boldsymbol \tau}$, we consider the  one-parameter velocity tensor:
 \begin{equation}
 \label{integration}
 \boldsymbol\nu_{\delta}(\mathbf r) = \Big\langle \mathbf v_{\boldsymbol\gamma(\delta)}(\mathbf r),\boldsymbol\gamma'(\delta)\Big\rangle.    
 \end{equation}
With this choice and setting $\phi_{0}(\mathbf r)=\mathbf r$, we obtain the deformation at the phase $\boldsymbol{\tau}$ as:
\begin{eqnarray}
\varphi_{\boldsymbol\tau}(\mathbf r) 
\label{straight} 
 &=&   \mathbf r + \int_{0}^{1} \Big\langle \mathbf v_{\boldsymbol\gamma(\delta)}\Big(\phi_{\delta}(\mathbf r)\Big),\boldsymbol\gamma'(\delta)\Big\rangle ~d\delta.
\end{eqnarray}

The above formulation produces parametric deformations $\varphi_{\boldsymbol\tau}$ that are dependent on the specific curve $\mathcal C_{\boldsymbol \tau}$. In this work, we define $\mathcal C_{\boldsymbol \tau}$ as a straight line segment that connects $\mathbf 0$ and $\boldsymbol{\tau}$. As discussed in Section \ref{background}, each of the deformation fields $\varphi_{\boldsymbol \tau}$ can be inverted by running the ODE backwards as in \eqref{odeback}. Thus, the image volume at phase $\tau_1$ can be deformed to phase $\tau_2$ using the transformation $\varphi_{\boldsymbol\tau_2} \circ ~\varphi_{\boldsymbol\tau_1}^{-1}$, where $\circ$ is the composition mapping.

\subsection{Path constraint on velocity tensors}
\label{velocity}
We note that $\varphi_{\boldsymbol{\tau}}$ in the previous section depends on the specific choice of the curve $C_{\boldsymbol{\tau}}$. We now consider a different path $\boldsymbol\beta:[0,1]\rightarrow\mathcal D_{\boldsymbol \tau}$ with $\mathcal D_{\boldsymbol\tau} \neq \mathcal C_{\boldsymbol\tau}$ that satisfies $\boldsymbol\beta(0) = \mathbf 0$ and $\boldsymbol\beta(1)=\boldsymbol{\tau}$. In this case, we obtain a diffeomorphism:
\begin{eqnarray}\label{crooked}
    \psi_{\boldsymbol{\tau}}(\mathbf r) &=&  \mathbf r + \int_{0}^{1} \Big\langle \mathbf v_{\boldsymbol\beta(\delta)}\Big(\psi_{\delta}(\mathbf r)\Big),\boldsymbol\beta'(\delta)\Big\rangle ~d\delta
\end{eqnarray}
With no additional constraints on the velocity tensor, we may have $\psi_{\boldsymbol{\tau}}\neq\varphi_{\boldsymbol{\tau}}$. If the integrals were path independent, a diffeomorphism could be directly evaluated from $\boldsymbol{\tau}_1$ to $\boldsymbol \tau_2$ by integrating the velocity tensor on a path connecting the two points in phase space rather than evaluating it as $\varphi_{\boldsymbol\tau_2} \circ \varphi_{\boldsymbol\tau_1}^{-1}$. 

We note that the constraint $\psi_{\tau}=\varphi_{\tau}$ is not necessary for the representation of the deformations and to realize the algorithm. However, in the interest of constraining the representation, we propose to add a penalty to encourage the diffeomorphisms to be path independent. We consider a penalty involving the path-dependent expected difference in the deformations $\mathbb E_{\mathcal D_{\tau}}\|\psi_{\tau}-\varphi_{\tau}\|^2$. In practice, at each epoch we choose $\mathcal D_{\boldsymbol{\tau}}$ as a random perturbation of the line segment $\mathcal C_{\boldsymbol{\tau}}$. We randomize the perturbations over epochs to ensure that the relation holds for all paths. Our in silico results in Fig. \ref{fig:regularization} show the benefit of adding this constraint. 

\subsection{Low-rank representation of velocity tensors}
\label{low-ranktensor}
The deformations $\varphi_{\boldsymbol \tau}(\mathbf r)$ are computed in terms of the velocity tensors $\mathbf v_{\boldsymbol \tau}(\mathbf r)$, parametrized by phase $\boldsymbol{\tau}$. There are extensive similarities between parametric tissue velocities $\mathbf v_{\boldsymbol{\tau}}$ in different phases. For example, deformations in different cardiac phases (for a specific respiratory phase) may be approximated by a relatively constant velocity tensor, as shown in Fig. \ref{fig:deform}. In the interest of realizing a compact representation, we propose to use a low-rank model to represent the velocity tensor at a specific phase $\boldsymbol{\tau}$:
\begin{equation}\label{lowrank}
    \mathbf v_{\boldsymbol{\tau}}(s,\mathbf r) = \sum_{n=0}^R \mathbf p_n(\mathbf r) ~m_n(s,\boldsymbol{\tau}) = \mathbf P \left(\mathbf{r}\right) \mathcal M_{\boldsymbol\kappa}(s,\boldsymbol\tau)
\end{equation}
Here $\mathbf{p}_n\left(\mathbf r\right): \mathbb R^3\rightarrow \mathbb R^2$ are the spatial basis functions, which are independent of $s$ and $\boldsymbol \tau$. $R$ denotes the rank of the representation. The weights $m_k(\boldsymbol{\tau})$ are nonlinear functions of $\boldsymbol{\tau}$ and $s$. We represent the non-linear function using a multilayer perceptron $\mathcal M_{\kappa}$ with parameters $\boldsymbol\kappa$.  The unknowns of the deformation model are $\boldsymbol\kappa$. Because the velocity tensors are spatially smooth, we represented $\mathbf p_n(\mathbf r)$ on a {$32^3$} grid, which is then interpolated to the $320^3$ volume using trilinear interpolation.

\section{D-MoCo implementation}
\begin{figure*}
\begin{center}
    \includegraphics[width=\textwidth, trim={8cm 7cm 8cm 7cm},clip]{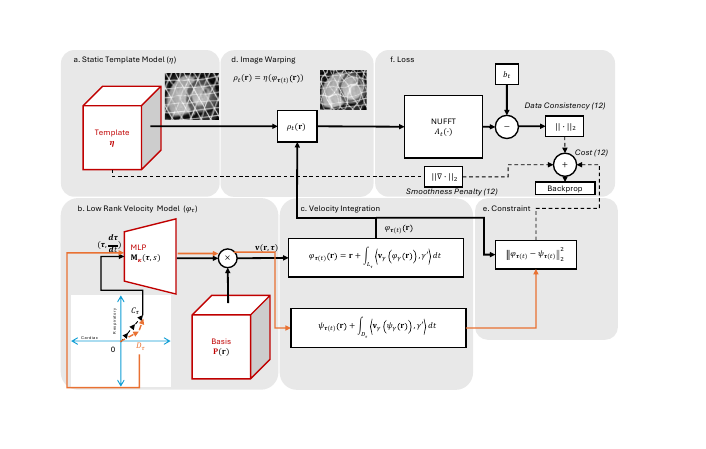}
\end{center}
    
 \vspace{-3.5em}    
    \caption{DMoCo architecture: (a) The static template volume $\boldsymbol\eta$.  (b) The velocity tensors at a specific phase $\boldsymbol{\tau}$ is modeled as the product of the static basis vectors $\mathbf P$ and its weights $m_n(s,\boldsymbol{\tau})$. The  weights are derived from the cardiac and respiratory phase $\boldsymbol{\tau} = (c,r)$ using a multilayer perceptron with parameters $\boldsymbol{\kappa}$.  (c) The deformation field are modeled as a forward diffeomorphic flow, denoted by \eqref{straight} along the straightline path $\mathcal C_{\boldsymbol \tau}$.  (d) The static image volume $\boldsymbol \eta$ is interpolated to each phase using trilinear interpolation. (e) The deformation is also derived along an alternate stochastic path $\mathcal D_{\boldsymbol \tau}$; the difference between the two different deformation estimates at $\boldsymbol \tau$ are used as a penalty, which is the second term in \eqref{fullopt}.  This term ensures that the velocity integrals are path independent, which further constraints the deformation estimates. (f) The forward operator involves multiplication by the coil sensitivity maps and NUFFT. Here, we consider $K$ readout groups with the closest phase to $\boldsymbol{\tau}$ and evaluate the $\ell_2$ error from the corresponding measurements measured k-space values.  A smoothness penalty is also added to the static template. }
    \label{Fig:art}
\end{figure*}

We estimate the static template $\boldsymbol{\eta}$ and the parameters of the deformation model from the measured k-space data in an unsupervised fashion. The preestimation of the motion phases is described in Section \ref{resp}. We formulate image recovery as an optimization scheme as described in \ref{fulloptsection}, where we estimate unknowns of the model: (a) low-rank factors $\mathbf P$ and MLP weights $\boldsymbol{\kappa}$ of the
deformation model in \eqref{lowrank} that the deformation model, and (b) the image template $\eta$. The fitting process is illustrated in Fig. \ref{Fig:art}. 

\subsection{Pre-estimation of the motion phases}
\label{resp}
We note that the data are acquired in a free-breathing and ungated fashion. The different readouts $b(t)$ that are acquired at different time points correspond to different cardiac and respiratory phases. We recorded the ECG triggers, from which we estimated the delay $c(t)$ of each radial spoke from the previous ECG trigger. We denote the maximum delay as $C = \max_t c(t)$. 

We estimate the respiratory phase from the SI k-space navigators using an auto-encoder. In particular, we reduce the SI signal to a one-dimensional latent variable, which captures the dominant signal variation in the navigator signal that results from respiratory motion. We denote the inverse Fourier transform of multichannel SI k-space navigators, denoted by $\mathbf x(t)$. We filter $x(t)$ with a temporal window of 1 Hz to remove cardiac and other variations in high-frequency signals.
We use a fully connected auto-encoder, with a one-dimensional latent vector $r(t)$. In particular, $r(t)$ is the one-dimensional vector that best explains the temporal variations in $x(t)$. The encoder $\mathcal E_{\theta_1}$ is an MLP with two hidden layers with 512 and 256 features, respectively. The decoder $\mathcal F_{\theta_2}$ uses the mirrored architecture, while the weights are not shared between the encoder and the decoder. The auto-encoder parameters are learned such that the loss
\begin{equation}
    L(\theta_1,\theta_2) = \sum_t\left\|\mathbf x(t)- \mathcal F_{\theta_1}\left(\underbrace{\mathcal E_{\theta_2} \left[\mathbf x(t)\right]}_{h(t)}\right)\right\|^2
\end{equation}
is minimized.

The MLP $\mathcal M_{\boldsymbol{\kappa}}$ in Section \ref{low-ranktensor} uses an input layer, which transforms the phase vector $\boldsymbol \tau = [c(t),r(t)]^T$ to the 3D vector $[\sin(2\pi c(t)/C) ,\cos(2\pi c(t)/C),r/R]$, where $R$ is the maximum magnitude of $r(t)$. This transformation preserves the complex harmonic nature of cardiac motion and the beat-to-beat variability of real-time imaging.

\subsection{Self-supervised Loss and Training}
\label{fulloptsection}
The unknowns of the proposed parametric deformable image model are
the static template $\eta$ in \eqref{imagemodel} and the parameters of the motion model $\varphi_{\boldsymbol\tau}$: the spatial basis functions $\mathbf P$ and the MLP parameters $\boldsymbol\kappa$.
We propose to recover these parameters by the minimization of the loss $L(\eta,\mathbf P, \boldsymbol \kappa)$:
\begin{eqnarray}\nonumber
    L&=&\sum_{\boldsymbol\tau}\Big(\overbrace{\left\|\mathcal A_{\tau} \Big(\eta(\varphi_{\boldsymbol \tau}(\mathbf r))\Big) - \mathbf b_{\boldsymbol \tau}\right\|^2}^{\rm data~consistency} \\\nonumber && +\overbrace{\lambda_1 ~\mathbb E_{\mathcal D_{\boldsymbol \tau}}\|\varphi_{\boldsymbol\tau}-\psi_{\boldsymbol\tau}\|^2}^{\rm velocity~constraint} \Big)\\\label{fullopt}
    &&  \qquad \qquad+
     \underbrace{\lambda_2 ~ \|\nabla \eta\|^2}_{\rm smoothness~penalty} 
\end{eqnarray}
The first term is the data consistency between the deformed version $\eta(\varphi_{\boldsymbol \tau}(\mathbf r))$ of the template $\eta(\mathbf r)$ at the phase $\boldsymbol{\tau}$ and the k-space data $\mathbf b_{\boldsymbol{\tau}}$ at that phase,  and the third is a spatial smoothness penalty on the static template. The second term in the above equation is a penalty to encourage the velocity tensors to be path-independent as described in Section \ref{velocity}. Here, $\varphi_{\boldsymbol{\tau}}$ is the deformation obtained by integrating the velocity tensor along a straight-line segment $\mathcal C_{\tau}$ as in \eqref{straight}. $\psi_{\boldsymbol{\tau}}$ is the integral along a perturbed path $\mathcal D_{\boldsymbol\tau}$ as in \eqref{crooked}. The second term involves an expectation of all possible paths between $0$ and $\boldsymbol \tau$. We evaluated both integrals using Euler's summation. The random path is obtained by perturbing the samples on the line by Gaussian noise of variance 10 percent of the step size between the time points. We use different random paths at each iteration/epoch of the optimization to encourage the deformation to be independent of the paths. 

We minimize the above loss using stochastic gradient descent. In each iteration, we pick a random $\boldsymbol{\tau} \in [0,1]^2$. We choose $K$ readout groups whose phases are the closest to $\tau$ in the Euclidean sense to define the loss and hence the gradient. In this work, we chose $K=20$. The hyperparameters are chosen by trial and error on one of the datasets and are kept fixed for the other datasets. We observe that the results were not very sensitive to $\lambda_1$, provided that it is large enough to ensure that the deformations are independent of the path. The parameter $\lambda_2$ was chosen to yield a good compromise between streak artifacts in the reconstructed template $\eta$ and blurring.

\subsection{Initialization of motion fields for fast convergence}
To speed up the training, we introduce a fast approach to initialize the velocity tensors. In particular, we group the k-space readouts to five cardiac and three respiratory motion bins, in an approach similar to XD-GRASP \cite{Feng2015}. We reconstructed the images corresponding to each of the bins using a smoothness regularized reconstruction:
\begin{equation}
\label{tikhonov}
\boldsymbol \rho = \arg \min_{\mathbf \rho}\sum_k\left\|A_{\boldsymbol\tau_k} \left(\boldsymbol \rho_{\boldsymbol \tau_k}\right)-\mathbf b_{\tau_k}\right|_2^2 + \lambda_3\left\|\nabla \boldsymbol \rho\right\|_2^2
\end{equation}
We note that each of the bins are heavily undersampled. We therefore chose a high value of $\lambda_3$ to obtain blurred images that are free of visible alias artifacts. We note that these images are only used to estimate the initial velocity tensors. We set the template as the image at $\boldsymbol{\tau}=0$ ($\eta = \boldsymbol \rho_{0}$) and solve for $\boldsymbol \tau$ and $\boldsymbol \kappa$ by minimizing the loss:
\begin{eqnarray}\nonumber\label{init}
    \mathcal L_{\rm init}\{ \mathbf P,\boldsymbol{\kappa}\} &=&\sum_{k} \left\|\eta\left(\phi_{\boldsymbol\tau_{k}}(\mathbf r)\right) - \boldsymbol \rho_{\boldsymbol{\tau_k}}\right\|^2 \nonumber\\ && + \lambda_1 \mathbb E_{\mathcal D_{\boldsymbol \tau}}\|\varphi_{\boldsymbol\tau}-\psi_{\boldsymbol\tau}\|^2
\end{eqnarray}
Because this optimization problem does not involve multichannel NUFFT, this initialization strategy is computationally much more efficient than that of \eqref{fullopt}. We use stochastic gradient descent to derive $\mathbf P$ and $\mathbf \kappa_{\rm init}$. We note that these parameters are used only to initialize the optimization scheme.

 \subsection{Image Acquisition}
The data are acquired using a 3D radial balanced steady-state free precession (bSSFP) research acquisition scheme using a spiral \textit{phyllotaxis} view ordering on a 1.5T MRI scanner (MAGNETOM Aera, Siemens Healthcare, Forchheim, Germany) system using a 32-channel cardiac array. The view order consisted of a superior-inferior (SI) k-space navigator, acquired in every 22 spokes. The temporal resolution of each group of 22 spokes is approximately $~$70 $\mathrm{ms}$. The number of radial readouts acquired per spoke was 5749, which corresponds to a fixed acquisition time of 5:58 minutes. The main acquisition parameters were as follows: TR/TE of 3.14/1.6 $\mathrm{ms}$, radio frequency excitation angle of $60^{\circ}$, resolution of 1.5 isotropic$\mathrm{mm}^3$, field of view (FOV) of 240 $\mathrm{mm}^3$, and readout bandwidth of 840 Hz/pixel \cite{Kettelkamp2023,Piccini2011}.

\subsection{State of the art (SOTA) methods used for comparison}
We compare our method against MoCo-SToRM \cite{moco-storm}. In MoCo-SToRM, each volume in the time series is modeled as the deformation of a static template, similar to the proposed approach. The main distinction is that the deformation fields are directly represented by a compact model in MoCo-STORM, unlike the low-rank representation of the velocity tensors that are integrated to obtain the deformation fields. To make the comparison fair, we represented the deformations in MoCo-STORM using a low-rank representation 
\begin{equation}
    \phi_{\tau}\left(\mathbf{r}\right) = \mathbf Q \left(\mathbf{r}\right) \mathcal N_{\boldsymbol\kappa}(\boldsymbol\tau).
\end{equation}
similar to \eqref{lowrank}.
Here, $\mathbf Q(\mathbf r)$ are the spatial basis vectors of dimension $R$ and $\mathcal N_{\kappa}$ is an MLP with parameters $\kappa$. 

We also implemented a motion-resolved approach (MoR) that is conceptually similar to \cite{Feng2015}. The cardiac data are grouped into three respiratory phases and five cardiac phases \cite{Feng2015}.  We note that with a short acquisition of 5:58 minutes, it is challenging to resolve more than five cardiac phases. The $\hat{\boldsymbol\rho}_{\boldsymbol \tau_{ij}}$ volumes are recovered using \eqref{eq::motion_resolved}.
\begin{eqnarray}
    \hat{\boldsymbol\rho}_{\boldsymbol\tau_k} &=& \arg \min_{\mathbf \rho}\sum_i\sum_j\left\|A_{\boldsymbol\tau_{ij}} \left(\boldsymbol \rho_{\boldsymbol \tau_{ij}}\right)-\mathbf b_{\tau_{ij}}\right\|_2^2 \nonumber\\ && + \lambda_4\left\|\nabla \boldsymbol  \rho\right\|_1
    \label{eq::motion_resolved}
\end{eqnarray}
Here we use the same notation as the above methods.  However, we use a temporal regularization between the bins $\left\|\nabla_{ij} \boldsymbol \rho\right\|_1$ denotes a total variation penalty.  The above motion-resolved reconstruction was done using a conjugate gradient algorithm. We also denote the variables as with the subscript $\tau_{ij}$ to show that the reconstruction is discretized to a fixed set of bins.

\section{Experimental setup}
 
\begin{figure*}
    \includegraphics[width=\textwidth, trim={0 4.5cm 3cm 0},clip]{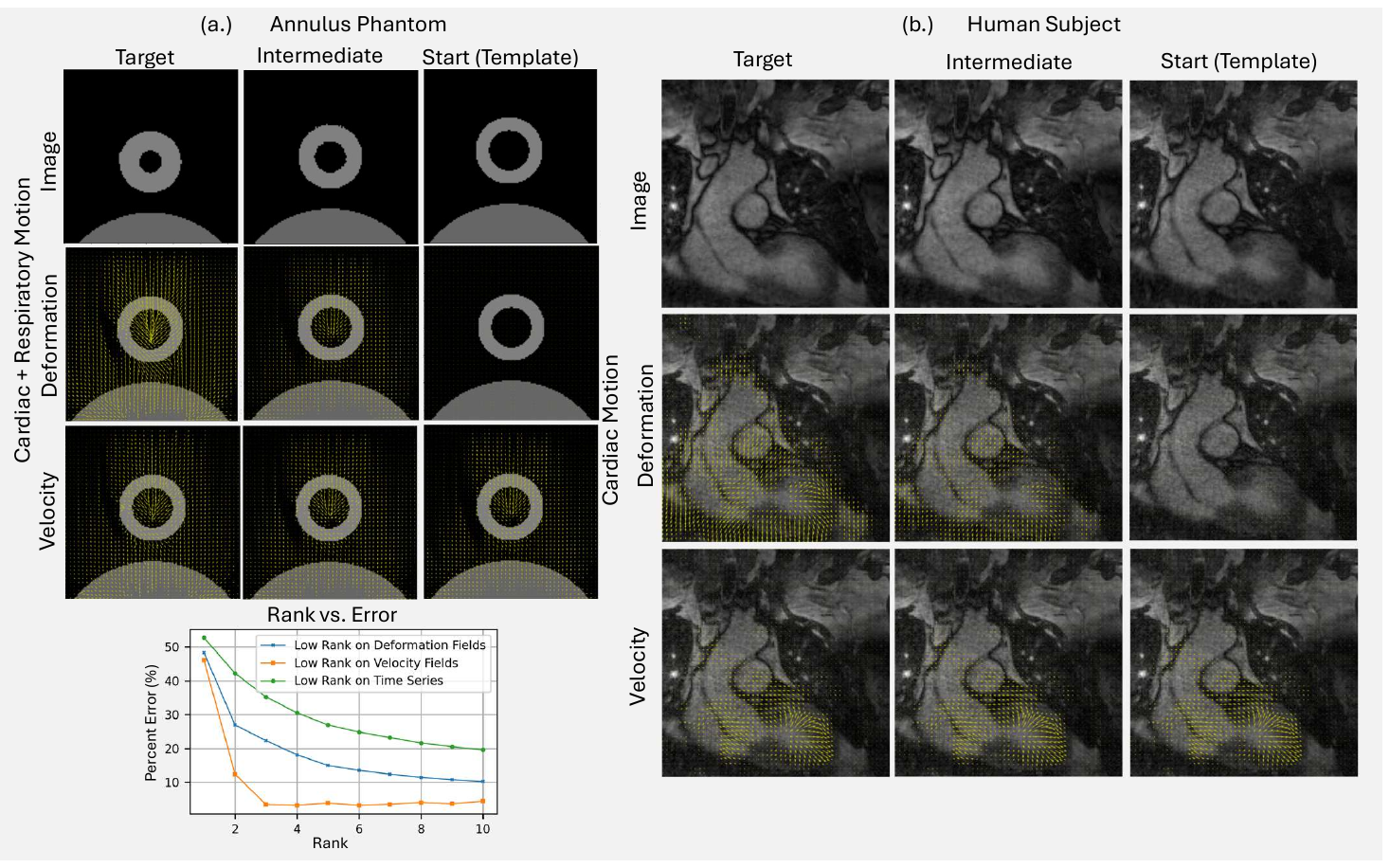}
    \caption{Impact of low-rank approximation of velocity tensors, deformation fields, and image series: We show the estimated velocity and deformation maps for a 2D annulus numerical phantom (top-left) and the 3D \textit{in-vivo} dataset (right). The top rows correspond to the images in three different phases, with the right-most image being the template. The middle rows correspond to the estimated deformations (shown by arrows), overlaid on the template, while the bottom rows show the velocities in these phases. We note that the velocity tensors (bottom rows) are almost the same across different phases, while the corresponding deformations are significantly more complex and changes drastically. In particular, the integration of a relatively static velocity tensor as in \eqref{straight} can account for complex time-varying deformation fields. In the plot (bottom-left) we performed a low-rank approximations with differing rank of the velocity tensors, deformation fields, and the image time series from the annulus phantom. We computed the corresponding reconstructions, which were compared against the original images.  The plots show the errors corresponding to low-rank approximations on the velocity tensors in orange, deformation fields in blue, and images in green. We observe that a significantly lower rank on the velocity tensors is sufficient to yield smaller errors, compared to the image time series and even the deformation fields. Here, a rank of 3 on the velocity tensor gave better results than a rank of 10 on the deformation fields.  These results show that a deformation model using low-rank modeling of velocities is very efficient in representing the image time series.}
    \label{fig:deform}
\end{figure*}

A challenge in quantitatively validating the proposed approach in an in-vivo setting is the lack of fully sampled cardiac and respiratory resolved 3D datasets; the acquisition of these data is impossible. We therefore restrict quantitative comparisons to a numerical phantom generated using XCAT simulations \cite{dukeXCATPhantom} and a 2D annulus with cardiac and respiratory motion Fig. \ref{fig:deform}. We also compare the proposed method with state-of-the-art methods using patient data sets. We note that quantitative image comparisons are challenging in this case due to drastically different tissue contrasts and mismatches in alignment. We instead visually compare 2D slices extracted from the reconstructed volumes that closely match with breath-held 2D cine acquisitions. We also performed segmentations of the left ventricles and perform a preliminary comparison with similar measures from 2D cine. 

\subsection{3D XCAT phantom}
We generated numerical data using the XCAT framework \cite{dukeXCATPhantom}, which consists of 5749 volumes, with respiratory and cardiac patterns. We assumed a respiratory period of 5 s and a heart rate of 60 beats per minute. We simulated the k-space measurements by multiplying the phantom by $B_1$ profile and simulated coil sensitivities of a birdcage coil
sensitivity mask with 6 coils, followed by a  non-uniform fast Fourier transformation (NUFFT) of each radial group. Noise of standard deviation $\sigma=5\times 10^{-3}$ was added to the k-space measurements. The image quality metrics were then calculated against the ground truth as well as the respiratory displacements (RD).  RD was assessed by measuring the displacement of the left diaphragm dome for the fitted volumes.   In addition, we compared the effect of modeling the flow and velocity tensors. The total simulated scan time was about six minutes and used the same FOV and resolution as in the \textit{ in vivo} study.

\subsection{2D Annulus phantom}
We use a 2D annulus phantom, considered in Fig. 2, to compare the low-rank approximations of the velocity tensors, deformation fields, and the image time series. The 2D setting makes the computational complexity and memory demand of the low-rank approximations manageable. Here, we consider an annulus that is expanding and contracting to mimic cardiac motion with 60 beats per minute. The cardiac motion was generated using two circles that contract at different rates. The radius of the outer circle decreases by 10\% of the FOV while the radius of the inner circle decreases by 14\% of the FOV.  The phantom also consists of a circular section at the bottom to mimic the liver. We also consider a contracting motion with a period of 4s and a displacement of 24\% 

 \subsection{Human subjects}
   The in-vivo data was acquired from five cardiac patients at the University of Iowa Hospitals and Clinics (UIHC), according to the institutional review board (IRB) protocols and the consent of the subject. The proposed sequence was added to the end of an existing clinical protocol, approximately 15 to 20 minutes after the short-acting Gadolinium (Gd) contrast agent was administered intravenously $0.1~mmol/kg$. 
   
   For qualitative comparisons, we extracted the time series of slices from the 3D reconstructions that closely match the 2D breath-held acquisitions with different views (e.g., short-axis, four-chamber, and two-chamber views). We note that separate breath-holds are required for the acquisition of each of the 2D stacks, while the proposed approach only involves a single free-breathing 3D acquisition. We note that the image contrast of the 3D acquisitions is drastically different from that of the 2D acquisitions, making quantitative image comparisons challenging.
    We compare the ejection fraction (EF), end-systolic volume (ESV), and end-diastolic volume (EDV)  against the values found in the 2D cine.
\begin{figure}
    \centering
    \includegraphics[width=0.5\textwidth, trim={0cm, 8.5cm, 17cm, 0cm}, clip]{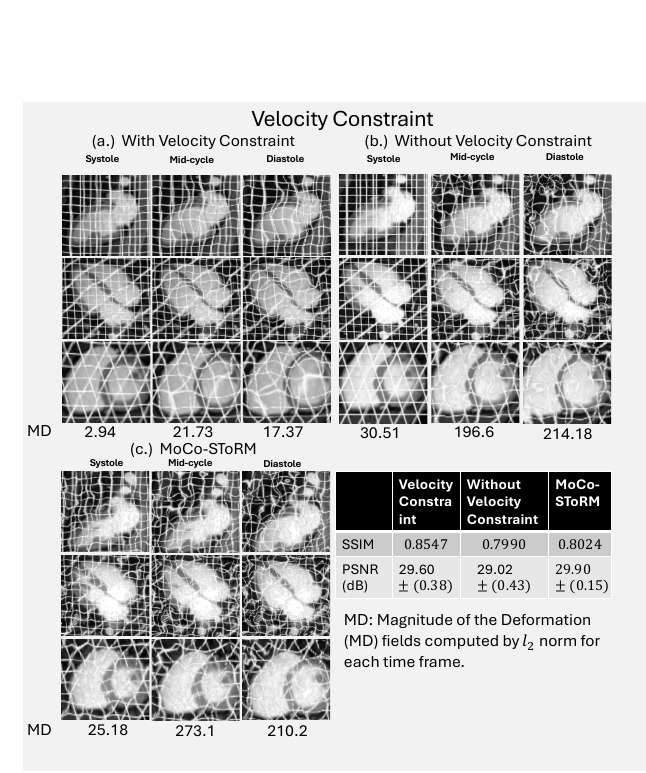}
    \caption{Impact of velocity constraint: We show the registered images and the deformation maps using (a) the proposed model, (b) the proposed model without the velocity constraint, and (c) the MoCo-SToRM method, which directly represents the deformation maps rather than integrating the velocities.  The  $l_2$ norm of the deformations is reported under each column denoted (MD); a higher number indicates more complex deformation. We also show the PSNR and SSIM measures of the moved images using all the methods, which show that all of them offer good registration. We note that registration is an ill-posed problem with multiple feasible deformations resulting in the same image. Our results show that the use of the velocity constraint translates to smoother deformations (lower MD and less wiggly deformations) than without the velocity constraint. We also note that the proposed DMoCo model is also more constrained than the MoCo-SToRM approach.
    }
    \label{fig:regularization}
\end{figure}

 \section{Results}
\subsection{\textit{Impact of low-rank approximation}}

\begin{figure}
    \centering
    \includegraphics[width=\linewidth, trim={0cm, 3.5cm, 9.5cm, 0cm}, clip]{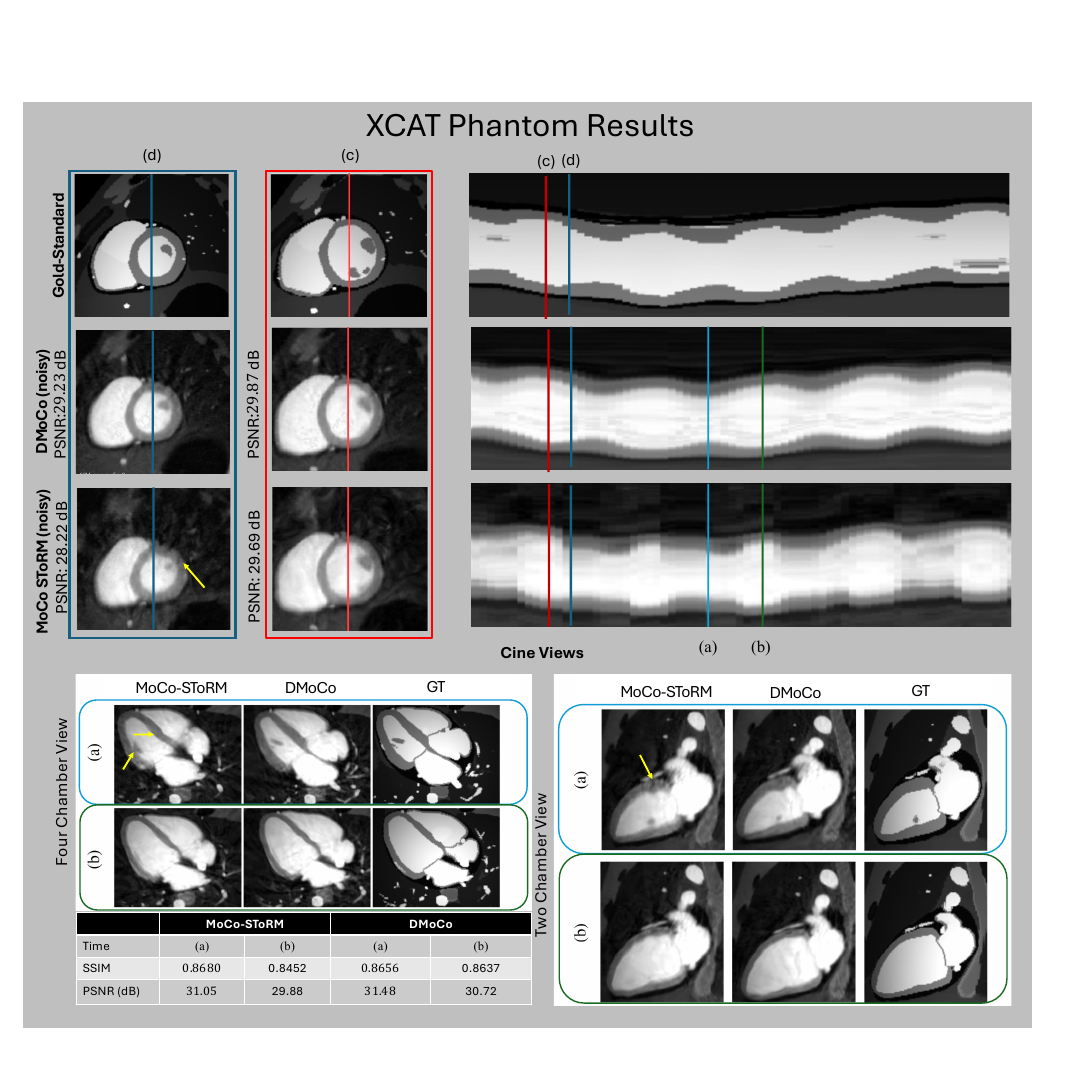}
    \caption{Comparison using XCAT phantom: We used the XCAT framework to simulate a free-breathing and ungated acquisition with the same settings as the in-vivo acquisition. We corrupted the simulated k-space observations with Gaussian noise of standard deviation $5 \times 10^{-3}$. We note that the acquisition scheme corresponds to a highly undersampling of several motion phases (e.g., systolic phases in end inspiration). The top three rows show a failure case, where the MoCo-SToRM approach resulted in a discontinuity in the recovered time profiles, shown on the right. This can also be appreciated from the recovered images (left panel with a blue border, corresponding to the blue lines in the time profiles on the right panel) in the corresponding phase, which shows increased blurring and a lower PSNR. By contrast, the DMoCo approach with a more constrained deformation model offers an improved reconstruction of these phases (see yellow arrows).  The bottom rows correspond to reconstructions from the systolic and diastolic phases, indicated by the light blue and green lines in the time profile. We observe increased noise blurring in the MoCo-SToRM reconstructions compared to DMoCo.}
    \label{fig:phantom}
\end{figure}
We first study the low-rank nature of the velocity and deformation tensors in Figure \ref{fig:deform} in a 2D annulus phantom in the left column (a).  The upper rows show three images at different physiological phases, with the right-most image as the template/source, which is in the systolic phase. The middle image corresponds to an intermediate phase. We perform a joint registration of volumes in the image domain using \eqref{init}. We overlay the deformation plots on the template in the middle row, while the velocity plots are overlaid in the bottom row. We observe that the velocity tensors are roughly similar across all phases, while the deformations are significantly different; the phase-dependent deformations can be obtained by integrating a relatively phase-independent velocity tensor according to \eqref{init}. We note that the integration in \eqref{integration} relies on the velocity tensors in the deformed state and therefore are non-linearly dependent on the velocity tensors. In the bottom row \ref{fig:deform}, we measure the distortion of the deformed images in each phase from the respective reference image as a function of the rank of velocity (orange), deformation fields (blue), and image series (green). Here, we use a 2D annulus with both respiratory and cardiac motion. We estimated the velocity and the corresponding deformations from the reference images using \eqref{init}. We then (a) truncated the velocity tensors to different ranks, and evaluated the corresponding deformation fields by integrating the approximated velocity tensors as in \eqref{integration}, followed by deforming the template to the respective phases using the corresponding deformation fields, (b) performed a low rank approximation of the deformation fields, followed by deforming the template to the respective phases using the low-rank approximations, and (c) performed a low-rank approximation of the images in the time series. We compare the approximate images to the actual ones to generate the error plot. 

The figures show that a velocity model of rank = 3 in the velocity is sufficient to produce low MSE values of 3. 45\%, while even a rank=10 approximation of the deformation fields and images can only give errors of 10.2\% and 22.3\%, respectively. The compact velocity representation is expected to yield more robust motion estimates, especially when estimated from noisy and undersampled images. This is considered in subsequent experiments. 

\subsection{\textit{Impact of velocity constraint}}
We study the impact of the velocity constraint in \eqref{fullopt} in Fig. \ref{fig:regularization}. We set the rank of the velocity model to ten. Here, we used \eqref{init} for the joint fit in the image domain, with the end systolic frame as the template. We show the estimated deformations for different cardiac phases, estimated using the proposed model with the velocity constraint in (a), without the velocity constraint in (b), and MoCo-SToRM in (c). We note that all models result in deformed images that closely match the original ones, as evidenced by the PSNR and SSIM metrics reported in the table in Fig. \ref{fig:regularization}. We note that there are numerous deformations of the template that can yield the targets. The results show that without the velocity constraint, the proposed scheme in (b) results in oscillatory deformations, indicated by the higher $\ell_2$ norm of the deformation (MD) shown below the images. The addition of the velocity constraint regularizes the deformations, resulting in deformations with significantly lower MD norm in (a). We also note that the deformations offered by MoCo-SToRM in (c), which directly models the deformation fields rather than the velocity tensors, result in deformations that are more complex than those of the proposed scheme. We note that the deformations are significantly more oscillatory in regions with low intensity away from the heart that are not seen from the above figures; these are not shown due to space constraints. These results show that DMoCo with velocity constraints offers a more constrained model that can account for motion while yielding smoother deformations. The resulting algorithm is thus expected to be less sensitive to undersampling artifacts and noise, which is studied in the next subsection.  

\begin{figure*}[!ht]
    \centering
    \includegraphics[width=0.8\textwidth, trim={0.0cm, 11.5cm, 19.5cm, 0cm}, clip]{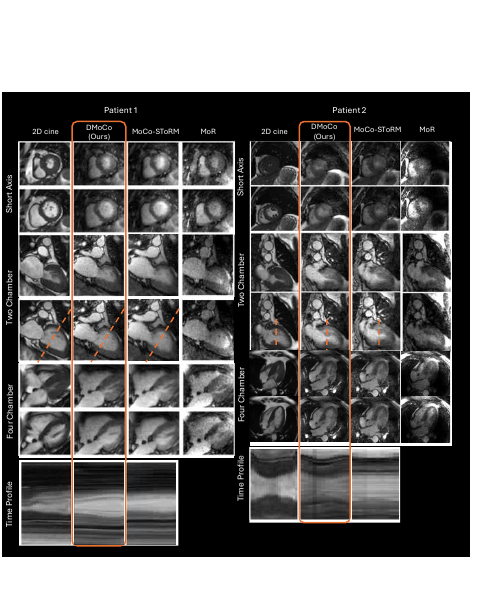}
    \caption{In-vivo experiments: We show the reconstructions obtained using DMoCO, MoCo-SToRM, and a motion resolved (MoR) approach top. We show the reconstructed short-axis and long-axis views and compare them against 2D cine. The top row of each view indicates ventricular systole, while the bottom corresponds to diastole. The bottom row shows the time profile of the red-dotted line in the two-chambered view.  The first column shows the 2D cine images, followed by our proposed method DMoCo, \cite{moco-storm}, and MoR.}
    \label{fig:in_vivo}
\end{figure*}

\subsection{\textit{Quantitative comparisons with SOTA methods.}}

We compare MoCo-SToRM and DMoCo on the X-CAT phantom in Fig. \ref{fig:phantom}, where the reconstruction is performed from noisy measurements in k-space. We considered the same sampling pattern as in the in-vivo experiments. We note from the time-profile plots that the MoCo-SToRM results in a piecewise-smooth profile with abrupt jumps, compared to the reference profiles.  We attribute the rapid jumps in the MoCo-SToRM temporal profiles to the less constrained deformation model. In contrast, the DMoCo approach yields smoother profiles, which are more consistent with the gold standard. The main distinction between the two approaches is the way in which the deformations are generated. The low-rank nature of the velocity tensors, which are integrated to obtain the deformations according to \eqref{integration}, as well as the velocity constraints within DMoCo encourage the deformations to be a smooth function of the cardiac phases. We also observe that the images recovered using DMoCo are less blurred and more consistent with the reference compared to MoCo-SToRM. Finally, we measure the respiratory displacement from the left dome of the diaphragm to be $ 26.06$ mm for DMoCo, $ 21.69$ mm for MoCo-SToRM,  compared to the ground truth of $ 27.33$ mm for the XCAT phantom, which also illustrates the improved accuracy of respiratory motion estimates. 

\subsection{\textit{In Vivo} Results}
The reconstructed images using the proposed DMoCo approach, MoCo-SToRM, and a motion resolved approach from two patients are shown in Fig. \ref{fig:in_vivo}. The first row of each patient corresponds to short-axis reconstructions in the systole, while the second row corresponds to the diastole. We also show results in the two-chamber and four-chamber views. The first column denotes the 2D SSFP cine acquisitions. We note that the contrast of the 2D cine acquisition is superior to that of the 3D acquisition. This is due to the inflow of unsaturated blood out of the slice to the slice, which appears significantly brighter than the saturated myocardium that experiences repeated radiofrequency excitation pulses. In contrast, the entire blood pool along with the myocardium is repeatedly excited in 3D acquisitions, translating to somewhat similar saturation levels. This results in a lower contrast-to-noise ratio between the blood pool and the myocardium. This is consistent with the literature \cite{inFlow}.

We note that the reconstructions with motion-resolved (MoR) approach appear more noisy than the motion-compensated (MoCo) schemes. As noted before, the motion-resolved schemes are sensitive to uneven sampling of the bins; the radial undersampling artifacts manifest as noise in the reconstructions. In contrast, motion-compensated methods combine the data from different motion states to recover the static template, which translates to reducing noise and artifacts. We do not show the temporal profiles of the MoR because it has only five cardiac phases. The results also show that the time profiles of the proposed model are more comparable to 2D cine, compared to MoCo-SToRM, which tends to have flatter time profiles.

\begin{figure*}[!ht]
    \centering
    \includegraphics[width=\textwidth, trim={0cm, 5cm, 0cm, 0cm}, clip]{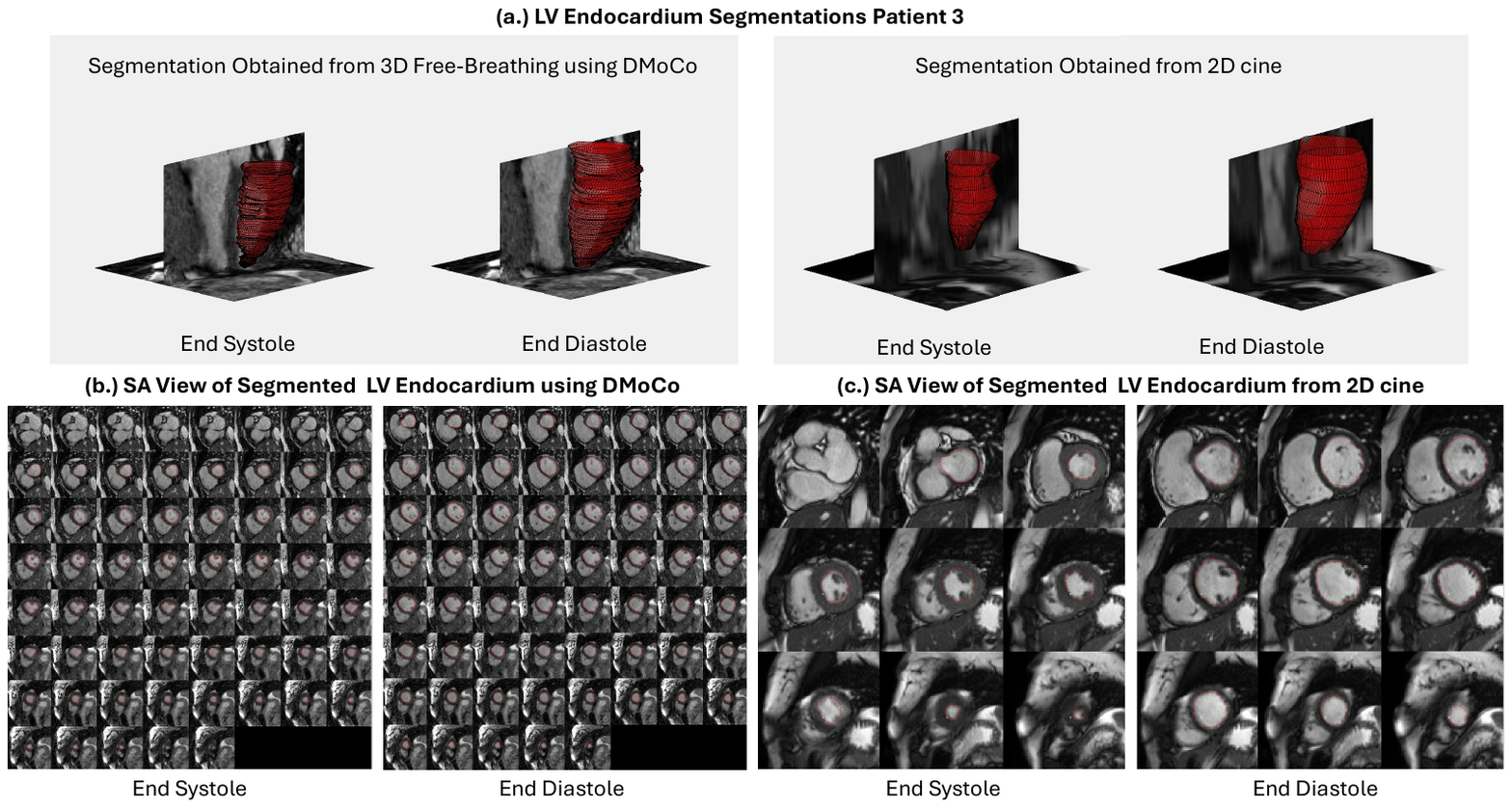}
    \caption{Reconstructed volumes and segmentations from patient three: We exported the short-axis projections of the 3D data from DMoCo reconstructions. The endocardium in each of the projections were segmented from both 2D cine and 3D DMoCo reconstructions using Segment-It, a free software from MEDVISE \cite{segmentit}.  (b) indicates the DMoCo reconstructions in end systole and diastole, together with the LV segmentations. (c) are the corresponding 2D BH cine slices, together with their segmentations. (a) is a 3D visualization if the segmented volumes using each method.   }
    \label{fig:segmentations}
\end{figure*}

\begin{figure*}[!ht]
    \centering
    \includegraphics[width=\textwidth, trim={0.5cm, 14.5cm, 12cm, 0cm}, clip]{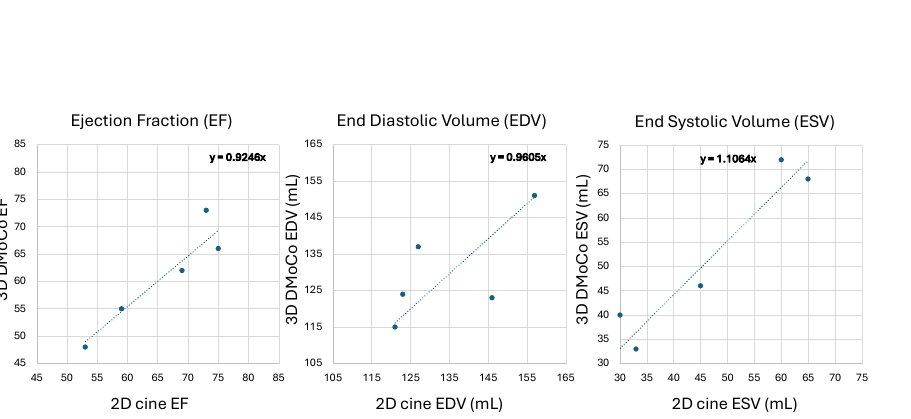}
    \caption{Comparison of functional measures with 2D BH cine acquisitions: We compared end-systolic, end-diastolic, and ejection fractions from DMoCo to 2D cine. We also show best fit line that passes through the origin.}
    \label{fig:baltanalysis}
\end{figure*}

 We segmented the left ventricles of the DMoCo reconstructions as shown in Fig. \ref{fig:segmentations}, which we compare against the 2D cine segmentations also shown. We compare the end diastolic and end systolic volumes, as well as the ejection fraction, in Fig. \ref{fig:baltanalysis}.  The results show moderate agreement between the 3D and 2D measures. We note that DMoCo offers isotropic resolution, whereas 2D cine offers highly anisotropic volumes and thick slices. These differences often translates to inconsistencies between the measures, which is also observed in several previous studies that compare functional measures between 3D and 2D \cite{2dv5dAlkassar2023,2dv5dGoo2018,Feng2017}. 
 
\section{Discussion}
The proposed DMoCo approach represents the images in the different motion phases as deformed versions of a static template. The deformation in each specific motion phase is represented as a diffeomorphism, obtained by integrating a parametric velocity tensor over intermediate phases, as shown in \eqref{straight}. The parametric velocity tensor itself is represented using a low-rank model, while a velocity constraint is added to encourage the deformations to be path independent. The parameters of the low-rank model and the static template are learned directly from the k-space data in a patient-specific, self-supervised fashion. 

The experiments on the numerical phantom show that the DMoCo model is more constrained than the MoCo-SToRM approach that uses a low-rank representation to directly model the deformations. Specifically, as illustrated in Fig. \ref{fig:deform}, the velocity model with a significantly lower rank can achieve high-quality deformations, compared to directly modeling the deformations or images using a low-rank representation. We also see from the results in Fig. \ref{fig:regularization} that the addition of the velocity constraints further regularizes the representation, while being able to accurately model the deformations. The integration of velocity tensors over the different intermediate phases to produce the deformation serves as a smoothness constraint between adjacent motion phases, compared to MoCo-SToRM. We see from Fig. \ref{fig:phantom} that the lack of smoothness constraints in MoCo-SToRM resulted in abrupt discontinuities in the time profiles and blurred reconstructions.  In contrast, the more constrained DMoCo deformation model yielded better preservation of the time profiles. This trend is also supported by more accurate measures of respiratory displacement (RD).

The experiments on in-vivo data are in line with the phantom results. We observe that the time profiles of DMoCo are more in agreement with the 2D cine acquisitions. The proposed scheme reduced the noise and alias artifacts compared to a motion-resolved reconstruction. The combination of information from different motion phases to reconsruct the template enables MoCo methods to be more resilient to undersampling and noise.  The ventricular volumes and ejection fractions measured from the 3D data are in moderate agreement  with those obtained from the 2D cine measurements, which is consistent with previous studies \cite{2dv5dAlkassar2023,2dv5dGoo2018,Feng2017}. A study in a larger cohort of patients is needed for a more thorough comparison.

A potential benefit of the proposed scheme is its ability to provide deformation maps. Although we have not studied the utility of these maps in estimating myocardial strain, this will be considered in the future. The focus of the present paper is to introduce the novel reconstruction algorithm and demonstrate its preliminary utility; the evaluation of the strain maps is beyond the scope of the present work. 

Despite promising results, the proposed study still has some limitations that restrict data quality, which we aim to overcome in our future studies. A challenge is the low contrast-to-noise ratio (CNR) of the 3D acquisitions. Unlike 2D scans that present good inflow contrast from unsaturated blood that flows into the imaging slice, the contrast in 3D acquisitions is inherently low. A possible alternative is a 3D multislab or interleaved multi-slice acquisition, which may offer improved contrast. Another challenge is that our current sequence did not employ fat suppression, resulting in strong residual streak artifacts. We will employ intermittent pulses of fat saturation as in \cite{Bastiaansen2019} to minimize residual fat signals. Furthermore, current studies show the exciting possibility of significantly improving CNR in 3D studies by using Ferumoxytol \cite{Whyferumoxy,Roy2022, Roy2024}, which may be another option to improve CNR.

The computational complexity of the current unsupervised approach is around 3 hours to reconstruct a dataset on an A100 GPU. While this is faster than the current motion resolved approaches, we believe that it can be accelerated in the future. We will consider more efficient variable splitting strategies or improved deep learning based priors (e.g. \cite{Chand2024}) on the template or initialization strategies, which may help manage the computational complexity. This will be the focus of our further research. The main focus of the current work is to introduce a novel method, and demonstrate its feasibility in a small group of subjects. A more elaborate study with more subjects is needed to fully evaluate the potential of this method in the clinic. 

\section{Conclusion}

An unsupervised motion-compensated image reconstruction algorithm was introduced for free-breathing and ungated 3D cardiac MRI. The images in different motion phases were expressed as the deformation of a static image template, where the deformations are represented by a low-rank diffeomorphic flow model. The static template and low-rank motion model parameters were learned directly from the k-space data in an unsupervised manner. The experiments show that the MoCo approach is less sensitive to undersampling of the motion phases. The more constrained motion model translated to more robust motion estimates, which offers improved performance compared to MoCo methods that directly model the deformations. 
\appendices

\bibliographystyle{unsrtnat}
\bibliography{tmi}%

\end{document}